\documentclass[letterpaper, 10 pt, conference]{ieeeconf}  

\IEEEoverridecommandlockouts                              

\makeatletter
\let\NAT@parse\undefined
\makeatother
\usepackage[numbers,compress]{natbib}

\usepackage{amsmath} 
\usepackage{amssymb}  
\usepackage{bm}
\usepackage{xcolor}
\usepackage{graphicx}
\usepackage{mathtools}
\usepackage{hyperref}
\usepackage{cleveref}
\usepackage{siunitx}
\usepackage{tikz}
\usetikzlibrary{shapes}
\usepackage{multirow}
\usepackage{makecell}
\usepackage{pifont}
\usepackage{cite}
\usepackage{subcaption}
\usepackage{tablefootnote}
\usepackage{eso-pic}
\newcommand{\copyrightnotice}{%
  \copyright\ 2025 IEEE. Personal use of this material is permitted. Permission from IEEE must be obtained for all other uses, in any current or future media,
  including reprinting/republishing this material for advertising or promotional purposes, creating new collective works, for resale or redistribution to
  servers or lists, or reuse of any copyrighted component of this work in other works.%
}

\Crefformat{figure}{#2Fig.~#1#3}
\Crefmultiformat{figure}{Figs.~#2#1#3}{ and~#2#1#3}{, #2#1#3}{ and~#2#1#3}
\Crefformat{equation}{#2Eq.~#1#3}

\begin{document}

\AddToShipoutPictureFG*{%
  \AtPageLowerLeft{%
    \put(\dimexpr (\paperwidth-\textwidth)/2 \relax, 25){%
      \parbox{\textwidth}{%
        \centering\footnotesize \copyrightnotice
      }%
    }%
  }%
}

\title{\LARGE \bf Whole-Body Inverse Dynamics MPC for Legged Loco-Manipulation}

\author{Lukas Molnar$^{1}$, Jin Cheng$^{2}$, Gabriele Fadini$^{2}$, Dongho Kang$^{2}$, Fatemeh Zargarbashi$^{2}$, Stelian Coros$^{2}$
\thanks{$^{1}$Lukas Molnar conducted this work as part of his Master's thesis in the Department of Mechanical and Process Engineering, ETH Zurich, Switzerland. He is now with Flexion Robotics \tt\footnotesize lukas@flexion.ai.}
\thanks{$^{2}$The remaining authors are with Computational Robotics Lab, Department of Computer Science, ETH Zurich, Switzerland. \{\tt\footnotesize jicheng, gfadini, kangd, fzargarbashi, scoros\}\tt\footnotesize @ethz.ch.}
\thanks{This work was supported by ETH Zurich RobotX Research Program.}
\thanks{Project website: \href{https://lukasmolnar.github.io/wb-mpc-locoman/}{\tt\footnotesize lukasmolnar.github.io/wb-mpc-locoman}}
\thanks{Code: \href{https://github.com/lukasmolnar/wb-mpc-locoman/}{\tt\footnotesize github.com/lukasmolnar/wb-mpc-locoman}}
}

\maketitle

\thispagestyle{empty}
\pagestyle{empty}

\begin{abstract}

Loco-manipulation demands coordinated whole-body motion to manipulate objects effectively while maintaining locomotion stability, presenting significant challenges for both planning and control. In this work, we propose a whole-body model predictive control (MPC) framework that directly optimizes joint torques through full-order inverse dynamics, enabling unified motion and force planning and execution within a single predictive layer. This approach allows emergent, physically consistent whole-body behaviors that account for the system’s dynamics and physical constraints.
We implement our MPC formulation using open software frameworks (\textit{Pinocchio} and \textit{CasADi}), along with the state-of-the-art interior-point solver \textit{Fatrop}. In real-world experiments on a \textit{Unitree B2} quadruped equipped with a \textit{Unitree Z1} manipulator arm, our MPC formulation achieves real-time performance at 80 Hz. We demonstrate loco-manipulation tasks that demand fine control over the end-effector’s position and force to perform real-world interactions like pulling heavy loads, pushing boxes, and wiping whiteboards.

\end{abstract}

\section{Introduction}

Legged robots equipped with robotic arms offer a powerful combination of mobility and manipulation capabilities, allowing them to interact with the environment while moving through it. To fully leverage this capability, such robots must execute precise end-effector tasks, as well as apply and resist substantial forces when pulling or pushing heavy objects, all while maintaining locomotion stability. These loco-manipulation tasks introduce strong coupling between the arm and the rest of the body, motivating control approaches that jointly optimize whole-body motion and force generation while enforcing physical constraints of the robotic system.

In this work, we address these challenges by developing a whole-body inverse dynamics model predictive control (MPC) framework that enables legged robots with arms to perform loco-manipulation tasks involving both precision and strong physical interaction. Unlike traditional methods for loco-manipulation that rely on reduced-order planning followed by whole-body tracking controllers, our approach directly optimizes whole-body dynamics at the torque level. This unified formulation enables simultaneous reasoning about motion and force generation while explicitly respecting actuation limits. As a result, we achieve emergent, dynamically consistent behaviors without relying on a hierarchical control stack, simplifying the overall pipeline.

We validate our approach on the \textit{Unitree B2} quadruped equipped with a \textit{Z1} manipulator arm, demonstrating real-time performance with the MPC solving at \SI{80}{\hertz}. In simulation, the controller enables accurate end-effector tracking across diverse whole-body motions and remains robust to external disturbances on the base. On hardware, the robot performs physically interactive tasks such as walking while pulling a \SI{10}{\kilogram} load, demonstrating the controller’s ability to coordinate locomotion and force-level manipulation under real-world conditions.

\begin{figure}
  \centering
  \begin{minipage}{\linewidth}
    \includegraphics[width=\linewidth]{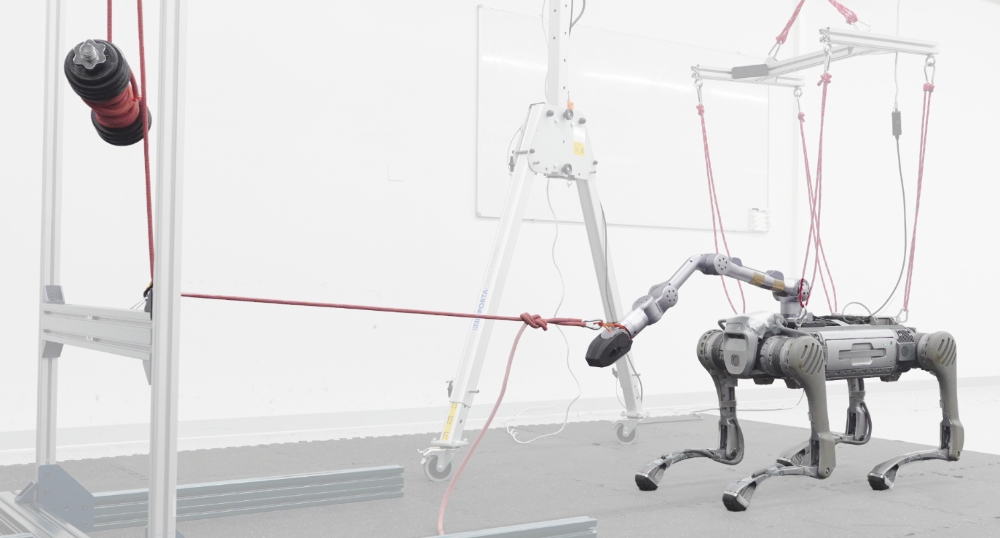}
  \end{minipage}
  \vspace{-2pt}
  \begin{minipage}{\linewidth}
      \includegraphics[width=0.44\linewidth]{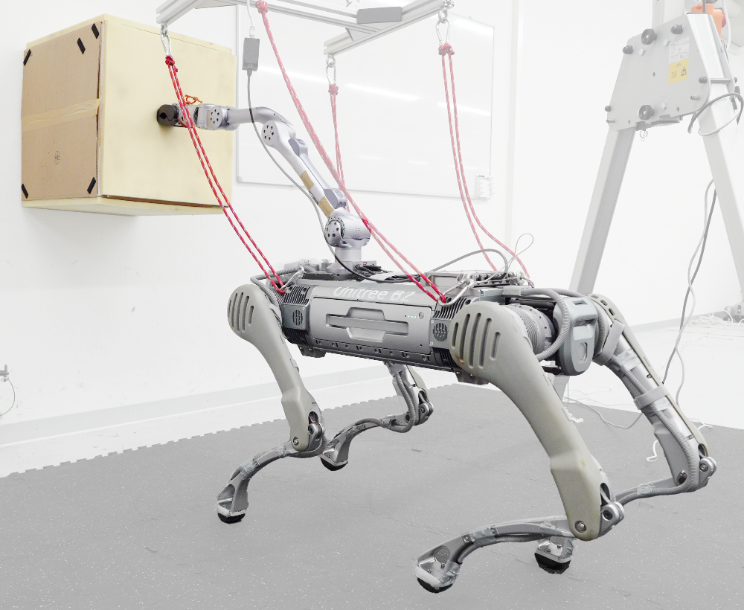}
      \includegraphics[width=0.55\linewidth]{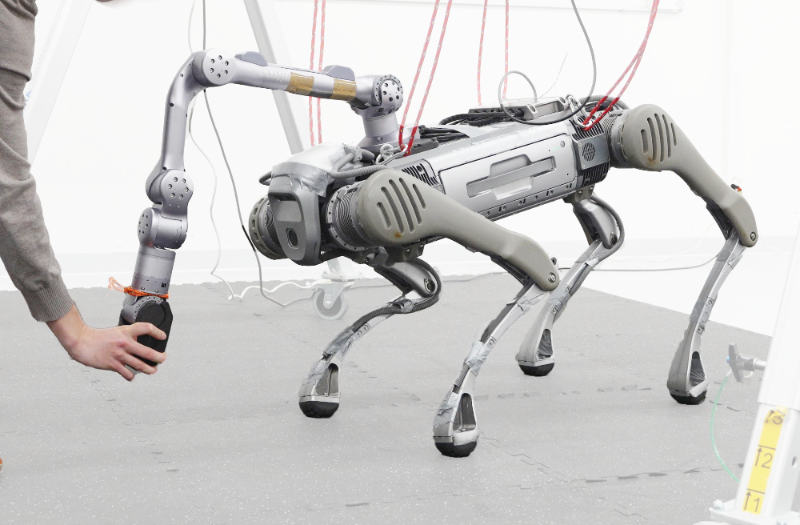}
  \end{minipage}
  \caption{\textit{Unitree B2} quadruped with a \textit{Z1} manipulator arm: Pulling a \SI{10}{\kilogram} load in stance \textbf{(top)}, pushing a box against the wall while walking \textbf{(bottom left)}, and demonstrating compliant human interaction \textbf{(bottom right)}.}
  \label{fig:hw-demo1}
  \vspace{-0.2cm}
\end{figure}

The main contributions of this work are as follows: 
\begin{enumerate}
    \item A whole-body torque-level MPC formulation based on inverse dynamics for legged loco-manipulation.
    \item A custom optimization framework built on \textit{\nobreak Pinocchio}~\citep{pinocchio} and \textit{CasADi}~\citep{casadi}, supporting multiple whole-body and centroidal dynamics formulations for benchmarking.
    \item Real-time hardware deployment using the interior-point solver \textit{Fatrop}~\citep{fatrop}, with the MPC solution applied directly to the robot without a low-level whole-body controller.
\end{enumerate}

\section{Related Work}

Control of legged robots has historically relied on a hierarchical structure that separates high-level planning from low-level control. At the highest level, many locomotion frameworks use a single rigid body (SRB) dynamics model to plan contact locations, contact forces, and a center-of-mass or base trajectory~\citep{mit-cheetah, mit-cheetah-wbc, anymal-srb, constrained-srb}. These models approximate the robot as a floating rigid body with constant inertia, ignoring joint-level kinematics, limb mass distribution, and actuation limits. To execute these plans on hardware, a low-level whole-body controller (WBC) with prioritized task execution is typically used~\citep{mit-cheetah-wbc, robust-wbc}, often relying on heuristics such as Bezier curves for swing leg trajectories.

Due to the computational efficiency of SRB models, they have recently been extended to loco-manipulation tasks~\citep{humanoid-locoman, hierarchical-locoman}. However, these methods require an additional higher-level planner to generate kinematically feasible references for the SRB model to track. This layered approach introduces hierarchical complexity, which requires tuning for each layer and can complicate deployment on hardware. Moreover, because SRB models assume massless limbs, they become inaccurate for systems with heavy manipulators, where the arm significantly affects the overall dynamics.

To improve consistency between planning and control, centroidal dynamics models have been introduced for quadruped and humanoid locomotion~\citep{centroidal-dyn-full-kin, rsl-dynamic-locomotion, perceptive-loco, centroidal-wb-comparison}. These models retain computational efficiency by focusing on the center of mass evolution, while incorporating the robot's full kinematic structure through the centroidal momentum matrix. This allows the planner to reason about joint motions and contact forces more accurately, enabling dynamic locomotion behaviors. For loco-manipulation, recent work has adopted centroidal dynamics for the ANYmal quadruped with a 4-DoF~\citep{unified-mpc} and 6-DoF manipulator~\citep{versatile-locoman, cheng2022haptic}, achieving dynamic tasks such as door opening and heavy object pushing. The same dynamics method is applied to the 37-DoF dual-arm quadruped CENTAURO~\citep{centroidal-37dof}, demonstrating the scalability of centroidal models for high-dimensional systems.

A key limitation of these methods is that they rely on additional steps to ensure dynamic feasibility at the joint level. For the ANYmal robot~\citep{unified-mpc, versatile-locoman}, a low-level whole-body controller based on hierarchical quadratic programming (QP) tracks the motions planned by the centroidal MPC, adding complexity to the control pipeline. In contrast, for the CENTAURO robot~\citep{centroidal-37dof}, joint torques are computed directly from the MPC solution via inverse dynamics, eliminating the need for a separate tracking controller. While this simplifies the control architecture, it assumes that the centroidal model produces motions consistent with the whole-body dynamics, which may result in unsafe torques in practice.

Whole-body torque-level MPC offers a unified optimization framework for simultaneous planning and control, and has shown success for both quadruped~\citep{rsl-wb-mpc, wb-impedance} and humanoid locomotion~\citep{wb-talos, wb-humanoid, mit-humanoid}. By directly modeling the full rigid-body dynamics and optimizing over joint torques, this approach enables physically consistent trajectory generation without requiring a separate low-level tracking controller. This is particularly appealing for loco-manipulation tasks involving intense environmental interactions---such as pushing or pulling heavy objects---where accurate modeling of force transmission and joint-level actuation is essential.

Despite the advantages, applying torque-level whole-body MPC to loco-manipulation remains largely unexplored, primarily due to the computational challenges of solving high-dimensional optimization problems in real-time. For pure locomotion, most real-time methods avoid optimizing contact forces explicitly, instead enforcing contact constraints within the dynamics~\citep{centroidal-wb-comparison, wb-impedance, wb-talos, wb-humanoid}. While this reduces the number of decision variables, it is not well-suited for manipulation tasks that require simultaneous control of end-effector force and motion---for example, when pulling a heavy object.

Full whole-body formulations that explicitly optimize both contact forces and joint torques~\citep{mit-humanoid, inv-dyn-mpc, rsl-wb-mpc} are better suited for loco-manipulation, but have received limited attention in this context. To date, such methods have only been applied offline for legged loco-manipulation~\citep{roloma}, without demonstrating real-time control. In this work, we address this gap by achieving real-time, torque-level whole-body MPC for loco-manipulation.

\section{Preliminaries}

\subsection{Model Predictive Control}
Model Predictive Control (MPC) is an optimization-based control strategy that computes actions by solving a finite-horizon optimal control problem (OCP) at each time step. The general form of the OCP for a discrete-time system can be expressed as:
\begin{equation}
\begin{aligned}
\min_{\boldsymbol{x}, \boldsymbol{u}} \quad & \sum_{k=0}^{N-1} \ell(\boldsymbol{x}_k, \boldsymbol{u}_k) + \ell_N(\boldsymbol{x}_N) \\
\text{s.t.} \quad & \boldsymbol{x}_0 = \hat{\boldsymbol{x}}, \\
& \boldsymbol{x}_{k+1} = \boldsymbol{f}(\boldsymbol{x}_k, \boldsymbol{u}_k), \quad \forall k = 0, \dots, N-1 \\
& \boldsymbol{g}(\boldsymbol{x}_k, \boldsymbol{u}_k) \leq 0, \quad \quad \forall k = 0, \dots, N-1 \\ 
& \boldsymbol{g}_N(\boldsymbol{x}_N) \leq 0
\end{aligned}
\end{equation}

Here, $\boldsymbol{x}_k$ and $\boldsymbol{u}_k$ denote the system state and control input at time step $k$, respectively. 
$\hat{\boldsymbol{x}}$ is the measured initial state, $\ell(\cdot)$ is the stage cost and $\ell_N(\cdot)$ the terminal cost.
The function $\boldsymbol{f}(\cdot)$ represents the system dynamics, 
$\boldsymbol{g}(\cdot)$ and $\boldsymbol{g}_N(\cdot)$  represent path constraints on each node, which may include both equality and inequality conditions.

\subsection{Whole-Body Dynamics}
We model the quadruped and manipulator system using a floating-base representation. The generalized coordinate vector is defined as $\boldsymbol{q} = [\boldsymbol{p}_b, \boldsymbol{q}_b, \boldsymbol{q}_j]$, consisting of the base position $\boldsymbol{p}_b \in \mathbb{R}^3$, base orientation $\boldsymbol{q}_b \in SO(3)$ represented as a unit quaternion, and joint angles $\boldsymbol{q}_j \in \mathbb{R}^{n_j}$. The corresponding generalized velocity and acceleration vectors are given by $\boldsymbol{v} \in \mathbb{R}^{6+n_j}, \boldsymbol{a} \in \mathbb{R}^{6+n_j}$, respectively, including linear and angular components for the base.

The full rigid-body (whole-body) dynamics are given by the following equations of motion:
\begin{equation}
\mathbf{M} (\boldsymbol{q}) \boldsymbol{a} + \mathbf{b} (\boldsymbol{q}, \boldsymbol{v}) =
\begin{bmatrix}
\mathbf{0}_{6 \times 1} \\
\boldsymbol{\tau}_j
\end{bmatrix}
+ \mathbf{J}_c^{\top}(\boldsymbol{q}) \boldsymbol{F}_c,
\label{eq:eom}
\end{equation}

where $\mathbf{M}(\boldsymbol{q}) \in \mathbb{R}^{(6 + n_j) \times (6 + n_j)}$ denotes the mass matrix, and $\mathbf{b}(\boldsymbol{q}, \boldsymbol{v}) \in \mathbb{R}^{6+n_j}$ represents the Coriolis and gravitational forces. The vector $\boldsymbol{F}_c = [\boldsymbol{F}_{c_1}^\top, \dots, \boldsymbol{F}_{c_4}^\top, \boldsymbol{F}_{c_\text{arm}}^\top]^\top \in \mathbb{R}^{18}$ concatenates the linear contact forces at the four feet and, for the arm end-effector, the linear force and optionally the angular moment. The term $\boldsymbol{\tau}_j \in \mathbb{R}^{n_j}$ denotes the joint torques, and $\mathbf{J}_c(\boldsymbol{q}) \in \mathbb{R}^{18 \times (6 + n_j)}$ is the contact Jacobian.

\section{MPC Formulation}

In this section, we present the MPC formulation for loco-manipulation using whole-body dynamics. The state vector $\boldsymbol{x} = [\boldsymbol{q}, \boldsymbol{v}]^{\top}$ consists of the generalized coordinates and velocities. The control input vector $\boldsymbol{u} = [\boldsymbol{\tau}_j, \boldsymbol{a}, \boldsymbol{F}_c]^{\top}$ includes joint torques, generalized accelerations, and contact forces.

\subsection{Cost Function}
\label{mpc-cost}
The stage cost is defined as a quadratic function that penalizes deviations from a desired state and control input:
\begin{equation}
\ell (\boldsymbol{x}_k, \boldsymbol{u}_k) = \left\| \boldsymbol{x}_k - \boldsymbol{x}^\text{des} \right\|_\mathbf{Q}^2
+ \left\| \boldsymbol{u}_k - \boldsymbol{u}^\text{des} \right\|_\mathbf{R}^2,
\end{equation}

where $\mathbf{Q} \succeq 0$ and $\mathbf{R} \succ 0$ are weight matrices for states and inputs, respectively. The terminal cost is chosen to be equivalent to the stage cost. The desired state and inputs are:
\begin{itemize}
    \item $\boldsymbol{q}^\text{des}$: the nominal robot configuration 
    \item $\boldsymbol{v}^\text{des}$: commanded linear and angular velocity of the base, along with zero joint velocities for regularization
    \item $\boldsymbol{a}^\text{des}, \boldsymbol{\tau}_j^\text{des}$: zero for regularization
    \item $\boldsymbol{F}_c^\text{des}$: evenly distribute the weight over the stance feet
\end{itemize}

\subsection{Constraints}
\subsubsection{Dynamics Constraints}
\label{sec:mpc-dyn}
To avoid defining quaternions as decision variables and enforcing their unit-norm constraint, we represent the generalized coordinates using a local increment $\delta \boldsymbol{q}_k \in \mathbb{R}^{6+n_j}$ in the tangent space of $SO(3)$. This increment is defined relative to the initial measured configuration $\hat{\boldsymbol{q}}$, and the absolute configuration is recovered via manifold integration, $\boldsymbol{q}_k = \hat{\boldsymbol{q}} \oplus \delta \boldsymbol{q}_k$. The state dynamics are discretized using explicit Euler integration with a time step $\delta t_k$:
\begin{equation}
\begin{aligned}
\delta \boldsymbol{q}_{k+1} &= \delta \boldsymbol{q}_k + \boldsymbol{v}_k \delta t_k, \\
\boldsymbol{v}_{k+1} &= \boldsymbol{v}_k + \boldsymbol{a}_k \delta t_k.
\end{aligned}
\label{eq:euler}
\end{equation}

The state dynamics in \Cref{eq:euler} represent a purely kinematic formulation. To ensure consistency with the whole-body dynamics, we impose the equations of motion (\Cref{eq:eom}) as path constraints at each time step $k$:
\begin{equation}
\begin{bmatrix}
\mathbf{0}_{6 \times 1} \\
\boldsymbol{\tau}_{j,k}
\end{bmatrix} = \mathbf{f}_\text{RNEA} \left( \boldsymbol{q}_k, \boldsymbol{v}_k, \boldsymbol{a}_k, \boldsymbol{F}_{c,k} \right).
\label{eq:rnea}
\end{equation}

Here, $\mathbf{f}_\text{RNEA}$ computes the inverse dynamics using the Recursive Newton-Euler Algorithm (RNEA)~\citep{rigid-body-dynamics}. Prior work has shown that direct shooting trajectory optimization problems, such as ours, benefit from using inverse dynamics rather than forward dynamics, resulting in faster solve times and greater robustness to coarse discretization~\citep{inv-vs-forw-dyn}. In~\Cref{sec:benchmark}, we validate this finding by benchmarking our inverse dynamics formulation against both forward dynamics and reduced-order centroidal dynamics.

\subsubsection{Contact and Swing Constraints}
The MPC receives a fixed gait schedule as an input, based on which the following contact and swing constraints are formulated:
\begin{equation}
\begin{aligned}
\text{Contact: } &
\begin{cases}
\boldsymbol{F}_{c_i,z} \geq 0, \quad\boldsymbol{v}_{c_i} = \boldsymbol{0}\\
\mu^2 \boldsymbol{F}_{c_i,z}^2 \geq \boldsymbol{F}_{c_i,x}^2 + \boldsymbol{F}_{c_i,y}^2. \\
\end{cases} \\
\text{Swing: } &
\boldsymbol{F}_{c_i} = \boldsymbol{0}, \quad
\boldsymbol{v}_{c_i,z} = \boldsymbol{v}_z^\text{ref}.
\end{aligned}
\end{equation}

The contact constraints ensure that the normal force $\boldsymbol{F}_{c_i,z}$ for foot $i$ is positive and inside the friction cone, with friction coefficient $\mu$. Furthermore, the linear velocity $\boldsymbol{v}_{c_i}$ of foot $i$ must be zero, ensuring the stance leg does not separate or slip with respect to the ground.

The swing constraints ensure that no external force acts on the swing feet and that the foot velocity tracks a reference trajectory in the vertical direction. This reference $\boldsymbol{v}_z^\text{ref}$ is parametrized by a cubic spline. In horizontal directions, the foot velocities are left unconstrained, allowing the MPC to optimize foot trajectories and step locations.

\subsubsection{Arm Constraints}
For the arm task, the desired end-effector force and velocity are constrained to track a user-defined command for all time steps $k$:
\begin{equation}
\begin{aligned}
\boldsymbol{F}_{c_\text{arm}} &= \boldsymbol{F}_{c_\text{arm}}^\text{des}, \quad
\boldsymbol{v}_{c_\text{arm}} = \boldsymbol{v}_{c_\text{arm}}^\text{des}.
\end{aligned}
\end{equation}

This formulation is suitable for tasks such as pulling a heavy object with a specified force and velocity.  However, it does not explicitly model the object or its dynamics, and therefore does not enable reasoning about how the object will respond to the applied forces.

\subsubsection{State and Input Bounds}
Joint positions and velocities are bound to their operational limits for all time steps $k$, while torque limits are only enforced for the first 2 time steps:
\begin{equation}
\begin{aligned}
\boldsymbol{q}_{j,\text{min}} &\leq \boldsymbol{q}_{j,k} \leq \boldsymbol{q}_{j,\text{max}} \quad \forall k,\\
\boldsymbol{v}_{j,\text{min}} &\leq \boldsymbol{v}_{j,k} \leq \boldsymbol{v}_{j,\text{max}} \quad \forall k,\\
\boldsymbol{\tau}_{j,\text{min}} &\leq \boldsymbol{\tau}_{j,k} \leq \boldsymbol{\tau}_{j,\text{max}} \quad \forall k \leq 2.
\end{aligned}
\end{equation}

Since the torques applied to the robot are interpolated only up to $k = 2$ (see~\Cref{sec:pipeline}), we remove torque limits beyond this node while ensuring safe hardware execution. This allows us to completely remove the torques from the decision variables for the rest of the horizon, while still guaranteeing dynamic feasibility through the first six rows of the RNEA constraint (\Cref{eq:rnea}). 

\subsection{Adaptive Time Steps}
\label{sec:adap-steps}
Accurate whole-body control requires high temporal resolution to capture fast dynamics, while the MPC must also look far enough ahead to plan footstep and joint trajectories effectively. Using uniformly small time steps across the entire horizon does not scale well computationally, as it requires a large number of nodes to span longer durations. To address this trade-off, previous work proposes a cascaded-fidelity MPC, using coarser time steps later in the horizon together with a simplified SRB dynamics model~\citep{cafe-mpc}. Other work on general MPC introduces an adaptive time step strategy that becomes exponentially more sparse throughout the horizon~\citep{diffusing-horizon-mpc}.

Inspired by these approaches, we adopt a geometric time step scheme of the form: $\delta t_k = \gamma^k \delta t_0$, with $\gamma > 1$. This provides fine resolution early in the horizon for accurate control, while reducing the number of nodes for long-term planning. The impact of this strategy compared to a uniform grid is evaluated in simulation (see \Cref{sec:res-sim-delay}). While our formulation uses whole-body dynamics throughout, incorporating a simplified model later in the horizon~\citep{cafe-mpc} may further improve scalability.

\section{Implementation Details}

We implement the MPC method using symbolic expressions of the system dynamics provided by the \textit{CasADi}~\citep{casadi} interface available in \textit{Pinocchio}~\citep{pinocchio}. \textit{CasADi} is additionally used to perform automatic differentiation and interface with numerical solvers. In this section, we provide implementation details of this integration and additional information related to the full deployment pipeline.

\subsection{Solution Method}
\label{sec:solution}
\subsubsection{Interior-Point Fatrop Solver}
We demonstrate that real-time whole-body MPC can be achieved using the interior-point (IP) solver \textit{Fatrop}~\citep{fatrop}, which solves the full nonlinear program (NLP) by leveraging the block-sparse structure of stage-wise constraints. This is enabled by a structure-exploiting linear solver based on Riccati recursion~\citep{fatrop-riccati}. In our experiments, \textit{Fatrop} achieves over a 10$\times$ speedup compared to the more standard IP solver \textit{IPOPT}~\citep{ipopt}, which does not explicitly exploit the problem structure. Its integration with \textit{CasADi} makes it easily applicable in our MPC pipeline. The main implementation requirement is that states and inputs must be interleaved in the decision variable vector to expose the stage-wise structure.

Most prior work on MPC for locomotion and loco-manipulation relies on sequential quadratic programming (SQP)~\citep{mit-humanoid, perceptive-loco, versatile-locoman}, which solves a sequence of quadratic subproblems obtained from successive linearizations of the nonlinear constraints. In contrast, \textit{Fatrop} allows us to solve the full nonlinear problem directly using second-order information from the exact Hessian. This approach achieves high accuracy, strict constraint satisfaction, and robust convergence with minimal parameter tuning, requiring as low as three iterations per solve.
\subsubsection{Warm-Starting}
Warm-starting solvers is generally critical for achieving real-time performance during hardware deployment. \textit{Fatrop} offers the option to be warm-started, which resulted in a \SIrange{15}{20}{\percent} reduction in solve time in our experiments. Each MPC iteration is warm-started using the solution from the previous step. However, the use of adaptive time steps introduces a slight time misalignment between successive horizons, which particularly affects the contact force variables due to mode switches between stance and swing phases. To mitigate this, we reinitialize the contact force guesses by distributing the robot’s weight evenly across stance feet and setting them to zero for swing feet.

\subsubsection{Code Generation}
Using \textit{Pinocchio}~\citep{pinocchio} and \textit{CasADi}~\citep{casadi}, the MPC can be implemented in Python and compiled into a shared library, which improves portability and facilitates fast execution. In our experiments, code generation resulted in a speed-up factor of 2 to 2.5. The compiled MPC can be directly evaluated for the \textit{Fatrop} solver, given the respective parameters and warm-started decision variables.
This enables straightforward hardware deployment, with the flexibility to tune parameters such as the state and input weights $\mathbf{Q}$ and $\mathbf{R}$ in real-time without recompiling the MPC.

\begin{figure}[t]
    \centering
    \includegraphics[width=\linewidth]{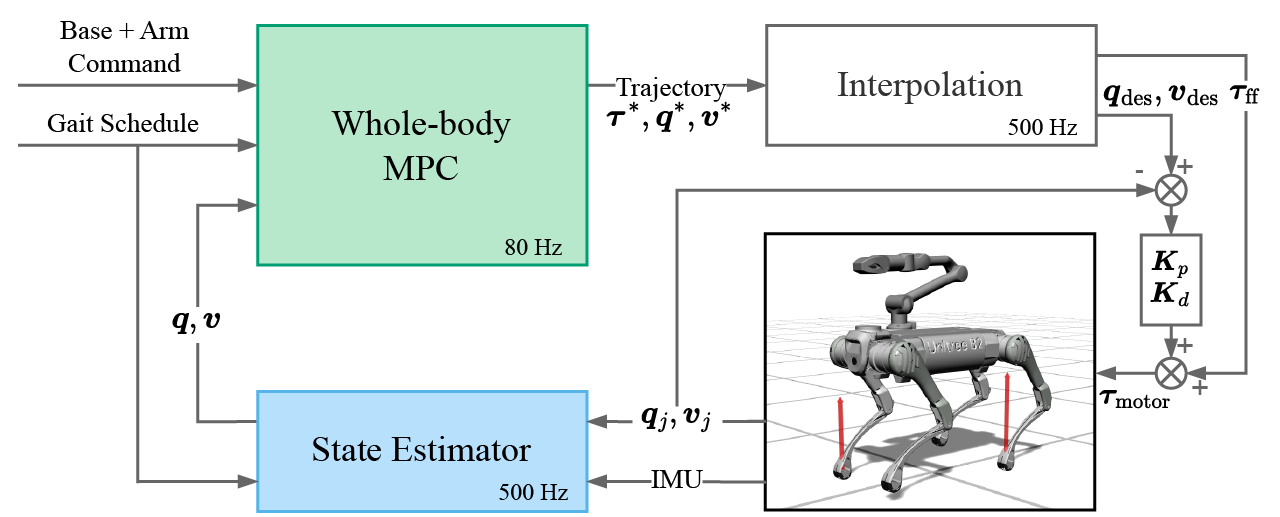}
    \caption{The pipeline of whole-body inverse dynamics MPC.} 
    \label{fig:pipeline}
    \vspace{-0.2cm}
\end{figure}

\subsection{Pipeline Description}
\label{sec:pipeline}

\Cref{fig:pipeline} displays the complete pipeline for deployment on the \textit{Unitree B2} quadruped, equipped with a \textit{Z1} manipulator. The \textit{B2} robot weighs approximately \SI{70}{\kilogram}, including the battery, and has three motors per leg, totaling 12 actuated joints. The \textit{Z1} robot weighs \SI{4.5}{\kilogram} and has 6 actuated joints. Since our experiments focus on linear arm tracking tasks, we lock the last 2 joints of the arm, resulting in a whole-body model with 22 degrees of freedom (6 base, 12 legs, 4 arm).

The MPC runs at \SI{80}{\hertz} on a PC equipped with an AMD Ryzen 9 9950X (16 cores, \SI{4.3}{\giga\hertz}). As inputs, it receives:
\begin{enumerate}
    \item \textbf{Tracking commands:} Targets for the base velocity (linear and angular), and arm end-effector velocity and force (linear).
    \item \textbf{Gait schedule:} Fixed contact schedule over the predicted horizon. The actual footstep locations are optimized by the MPC.
    \item \textbf{Initial state:} Generalized coordinate positions and velocities from the state estimator.
\end{enumerate}

The MPC outputs desired joint position, velocity, and torque trajectories. These are interpolated linearly at \SI{500}{\hertz}:
\begin{equation}
\begin{aligned}
\boldsymbol{\tau}_\text{ff}(t) &= \boldsymbol{\tau}_0^* + \frac{\boldsymbol{\tau}_1^* - \boldsymbol{\tau}_0^*}{\delta t_0} (t - t_0), \\
\boldsymbol{q}_\text{des}(t) &= \boldsymbol{q}_0^* + \boldsymbol{v}_0^* (t - t_0), \\
{\boldsymbol{v}}_\text{des}(t) &= \boldsymbol{v}_0^* + \boldsymbol{a}_0^* (t - t_0).
\end{aligned}
\end{equation}

\begin{table*}[t]
\setlength{\tabcolsep}{5pt}
\centering
\scriptsize
\begin{tabular}{|c|c|c|c|c|c|c|c|c|c|}
\hline
\rule{0pt}{3.5ex}
\multirow{2}{*}{\textbf{Model}} &
\multirow{2}{*}{\makecell{\textbf{Related} \\ \textbf{Work}}} &
\multirow{2}{*}{\textbf{States}} &
\multirow{2}{*}{\textbf{Inputs}} &
\multirow{2}{*}{\textbf{Dynamics Constraint}} &
\multirow{2}{*}{\makecell{\textbf{Path} \\ \textbf{Constr.}}} &
\multirow{2}{*}{\makecell{\textbf{Instr. per} \\ \textbf{Constr.}}} &
\multirow{2}{*}{\makecell{\textbf{Dec.} \\ \textbf{Vars.}}} &
\multicolumn{2}{c|}{\makecell{\textbf{Solve Time [ms]} \\ \textbf{avg (std)}}} \\
\cline{9-10}
\rule{0pt}{2ex}
&&&&&&&& \textbf{Pre-comp.} & \textbf{Post-comp.} \\
\hline

\rule{0pt}{4ex}
\makecell{Whole-Body \\ Inv. Dyn. (\textbf{Ours})} &
\citep{mit-humanoid, inv-dyn-mpc, roloma} &
$\begin{bmatrix} \boldsymbol{q}, \boldsymbol{v} \end{bmatrix}$ &
$\begin{bmatrix} \boldsymbol{\tau}_j, \boldsymbol{a}, \boldsymbol{F}_c \end{bmatrix}$ & 
$\begin{bmatrix}
\mathbf{0}_{6 \times 1} \\ \boldsymbol{\tau}_j
\end{bmatrix} =
\mathbf{M} \boldsymbol{a} + \mathbf{b} - \mathbf{J}_c^{\top} \boldsymbol{F}_c$ &
\ding{51} & \textbf{3818} & 1226 & \textbf{29.0} (0.9) & \textbf{12.5} (0.5) \\ [2ex]
\hline

\rule{0pt}{4ex}
\makecell{Whole-Body \\ Forw. Dyn.} &
\citep{rsl-wb-mpc, wb-mobile-manipulator} &
$\begin{bmatrix} \boldsymbol{q}, \boldsymbol{v} \end{bmatrix}$ &
$\begin{bmatrix} \boldsymbol{\tau}_j, \boldsymbol{F}_c \end{bmatrix}$ & 
$\boldsymbol{a} = \mathbf{M}^{-1} \left(
\begin{bmatrix}
\mathbf{0}_{6 \times 1} \\ \boldsymbol{\tau}_j
\end{bmatrix}
+ \mathbf{J}_c^{\top} \boldsymbol{F}_c - \mathbf{b}
\right)$ &
\ding{55} & 7764 & 1094 & 81.6 (9.2) & 34.2 (3.8) \\ [2ex]
\hline

\rule{0pt}{4ex}
\multirow{2}{*}{\makecell{
\rule{0pt}{4ex}
Centroidal \\ Velocity}} &
\citep{centroidal-dyn-full-kin} &
$\begin{bmatrix} \boldsymbol{h}_\text{com}, \boldsymbol{q} \end{bmatrix}$ &
$\begin{bmatrix} \boldsymbol{v}, \boldsymbol{F}_c \end{bmatrix}$ &
$\boldsymbol{h}_\text{com} = \mathbf{A} \boldsymbol{v} =
\begin{bmatrix}
    \mathbf{A}_b \ \mathbf{A}_j
\end{bmatrix}
\begin{bmatrix}
    \boldsymbol{v}_b \\ \boldsymbol{v}_j
\end{bmatrix}$ &
\ding{51} & 4070 & 938 & 33.0 (1.9) & 16.0 (0.9) \\ [2ex]
\cline{2-10}

\rule{0pt}{4ex}
& \citep{unified-mpc, versatile-locoman, cheng2022haptic, centroidal-37dof} &
$\begin{bmatrix} \boldsymbol{h}_\text{com}, \boldsymbol{q} \end{bmatrix}$ &
$\begin{bmatrix} \boldsymbol{v}_j, \boldsymbol{F}_c \end{bmatrix}$ & 
$\boldsymbol{v}_b = \mathbf{A}_b^{-1} \left( \boldsymbol{h}_\text{com} - \mathbf{A}_j \boldsymbol{v}_j \right)$ &
\ding{55} & 4727 & \textbf{854} & 115.5 (14.5) & --- * \\ [2ex]
\hline

\rule{0pt}{4ex}
\multirow{2}{*}{\makecell{
\rule{0pt}{4ex}
Centroidal \\ Acceleration}} &
--- &
$\begin{bmatrix} \boldsymbol{q}, \boldsymbol{v} \end{bmatrix}$ &
$\begin{bmatrix} \boldsymbol{a}, \boldsymbol{F}_c \end{bmatrix}$ & 
$\dot{\boldsymbol{h}}_\text{com} = \dot{\mathbf{A}} \boldsymbol{v} + 
\begin{bmatrix}
    \mathbf{A}_b \ \mathbf{A}_j
\end{bmatrix}
\begin{bmatrix}
    \boldsymbol{a}_b \\ \boldsymbol{a}_j
\end{bmatrix}$ &
\ding{51} & 15628 & 1178 & 103.7 (3.9) & --- * \\ [2ex]
\cline{2-10}

\rule{0pt}{4ex}
& \citep{centroidal-wb-comparison, aligator} &
$\begin{bmatrix} \boldsymbol{q}, \boldsymbol{v} \end{bmatrix}$ &
$\begin{bmatrix} \boldsymbol{a}_j, \boldsymbol{F}_c \end{bmatrix}$ & 
$\boldsymbol{a}_b = \mathbf{A}_b^{-1} \left( \dot{\boldsymbol{h}}_\text{com} - \dot{\mathbf{A}} \boldsymbol{v} - \mathbf{A}_j \boldsymbol{a}_j \right)$ &
\ding{55} & 16339 & 1094 & 189.8 (18.0) & --- * \\ [2ex]
\hline

\end{tabular}
\caption{A comparison of our whole-body inverse dynamics MPC formulation with forward dynamics and reduced-order centroidal models, all of which are available in our framework. We evaluate solve times before and after generating C code for the OCP and compiling it to a shared library for hardware transfer. (*) Indicates that compiling the OCP failed due to memory exhaustion, on a system with \SI{32}{\giga\byte} RAM.}
\label{tab:benchmark}
\vspace{-0.2cm}
\end{table*}

If $t_1$ has passed and we have not yet received a new solution from the MPC module, we interpolate the last received solution analogously between $t_1$ and $t_2$. The interpolated commands are then fed into a low-level PD controller to compute the desired motor torque:
\begin{equation}
\boldsymbol{\tau}_\text{motor} = \boldsymbol{\tau}_\text{ff} + \boldsymbol{K}_p (\boldsymbol{q}_\text{des} - \boldsymbol{q}) + \boldsymbol{K}_d (\boldsymbol{v}_\text{des} - \boldsymbol{v}).
\end{equation}

This joint-level feedback is only required for hardware deployment and is standard practice to compensate for actuator limitations and unmodeled dynamics, particularly in systems with quasi-direct drive actuators.

Finally, IMU and joint encoder readings are processed by a state estimator to update the initial state of the MPC. As the \textit{B2} robot does not provide contact sensing, the estimator infers foot contact states from the planned gait schedule, following a two-stage Kalman filter approach~\citep{mit-cheetah-3}. Furthermore, since the joint encoder velocities seemed to be noisy on hardware, an additional low-pass filter is added at the input to the MPC, with a cutoff frequency of \SI{50}{\hertz}.

\section{Results}

We first benchmark our whole-body inverse dynamics formulation against other state-of-the-art modeling approaches, highlighting the computational efficiency of our method. Following this, we evaluate the precise and robust loco-manipulation behaviors of our MPC in simulation. Finally, we demonstrate its capabilities on hardware through a series of interactive loco-manipulation tasks.

\subsection{Benchmarking Dynamics Models}
\label{sec:benchmark}

In~\Cref{tab:benchmark}, we compare our whole-body inverse dynamics MPC against forward dynamics, as well as reduced-order centroidal models, the latter representing the current state-of-the-art for loco-manipulation \citep{unified-mpc, versatile-locoman}. All dynamics formulations are implemented in our proposed framework, allowing for systematic comparisons. 

The centroidal methods model the evolution of the center of mass (CoM) linear and angular momentum $\boldsymbol{h}_\text{com} \in \mathbb{R}^6$. They describe the relationship of the CoM with the kinematic configuration of the robot through the centroidal momentum matrix (CMM) $\mathbf{A}(\boldsymbol{q})$. For both velocity- and acceleration-level centroidal dynamics, we can add the resulting CMM constraint as a path constraint, or by propagating it through the state transition function $\boldsymbol{f}(\cdot)$ (see~\Cref{tab:benchmark}). The latter case is employed if we remove the base velocity/acceleration from the input, which is the more common choice in practice since the base is unactuated.

We compare the solve times for all dynamics formulations using the \textit{B2+Z1} system. The MPC is evaluated in closed loop using Pinocchio’s forward dynamics without a high-fidelity physics simulator. It contains 15 nodes with adaptive time steps, covering a \SI{0.56}{\second} horizon (\SI{70}{\percent} of the gait period). We found that solve times for \textit{Fatrop} scale approximately linearly with the number of nodes, and this choice provided a good balance between accuracy and computational speed. To match the constraints of the whole-body MPC, torque limits are applied to the centroidal formulations for $k \leq 2$, by estimating them through the contact Jacobian and gravity compensation.

The results in ~\Cref{tab:benchmark} show that our whole-body inverse dynamics MPC achieves the fastest solve times among all formulations, despite having the highest number of decision variables. It solves the nonlinear optimization problem to convergence in \SI{12.5}{\milli\second} post-compilation, enabling real-time control at \SI{80}{\hertz}. To help interpret this result, we evaluate the number of mathematical instructions in the \textit{CasADi} expression for each constraint. We see that the inverse dynamics RNEA constraint has the lowest number of instructions, likely explaining the computational speed.

Furthermore, \textit{Fatrop} consistently performs far better when dynamics are imposed as path constraints rather than embedded in the state transition function. Although both optimization formulations use direct multiple shooting, embedding the dynamics in $\boldsymbol{f}(\cdot)$ effectively forward-propagates the input initial guess, which can drive the state trajectory farther from the optimum. In contrast, enforcing the dynamics as path constraints allows the solver to jointly optimize over states and inputs, resulting in faster convergence with lower variance (see ~\Cref{tab:benchmark}).

\subsection{Simulation Experiments}
\label{sec:res-sim}
We run our whole-body MPC for the \textit{B2+Z1} model in a physics simulator based on the Open Dynamics Engine (ODE)~\citep{ode}. We evaluate the controller's performance with and without accounting for the computation time delay.

\begin{figure}[t]
    \centering
    \vspace{0.05cm}
    \includegraphics[width=\linewidth]{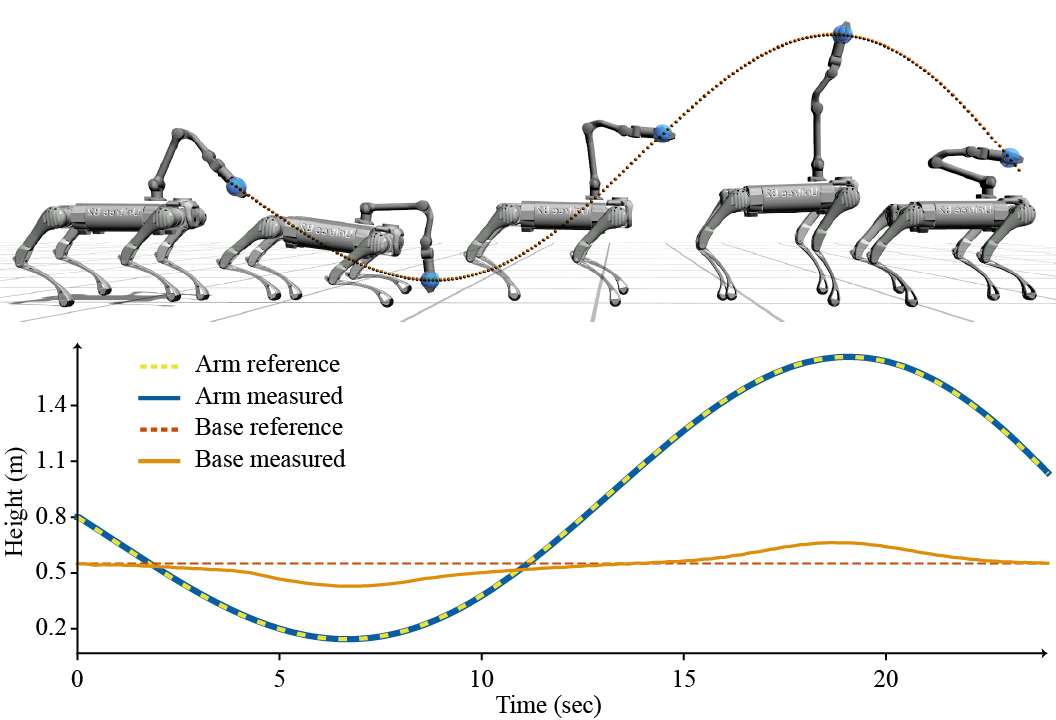}
    \caption{Whole-body adaptability: The MPC generates coordinated full-body motions to track an arm reference trajectory while trotting, automatically adjusting the base pose to reach both low and high targets.}
    \label{fig:wb-adaptability}
    \vspace{-0.2cm}
\end{figure}

\subsubsection{Without Time Delay}
Without accounting for the computation time delay, our MPC achieves stable locomotion and end-effector tracking with pure feed-forward torques ($\boldsymbol{K}_p = \boldsymbol{K}_d = 0$). We perform several experiments to test the capabilities of the controller.

\Cref{fig:wb-adaptability} illustrates adaptive whole-body behavior across varying manipulator targets. In this experiment, the robot trots forward at \SI{0.2}{\meter/\second} while tracking a reference trajectory with the arm end-effector. The trajectory is converted into velocity commands using position feedback, enabling highly accurate tracking as visible in~\Cref{fig:wb-adaptability}. Furthermore, when reaching down low or up high, the MPC automatically adjusts the base pose and leg motions to maintain tracking accuracy for the manipulator. This behavior emerges from the cost design: base deviations are penalized softly, while the end-effector velocity is enforced as a hard constraint. These results demonstrate the ability of the whole-body MPC to achieve precise arm tracking by optimizing base and joint motions in coordination.

\begin{figure}[t]
    \centering
    \includegraphics[width=\linewidth]{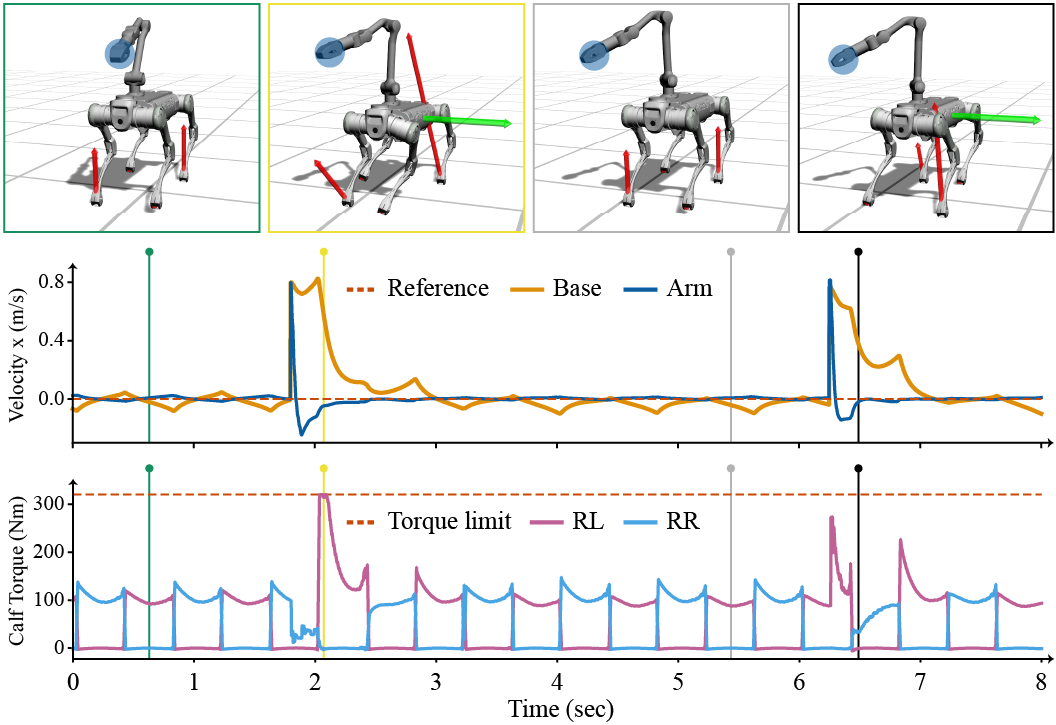}
    \caption{Disturbance rejection: The base velocity is perturbed laterally by \SI{0.8}{\meter/\second} while trotting in place. The arm end-effector and base lateral velocities are visualized, along with the rear leg calf torques (RL and RR).}
    \label{fig:disturbance}
    \vspace{-0.2cm}
\end{figure}

\Cref{fig:disturbance} demonstrates the second simulation experiment, in which a velocity disturbance of \SI{0.8}{\meter/\second} is applied to the robot's base while trotting in place. The MPC successfully stabilizes this disturbance while maintaining high end-effector tracking accuracy. Additionally, the joint torques always stay within the operational limits, once even pushing the rear left calf joint to its \SI{320}{\newton\meter} limit. This illustrates the benefit of our torque-level MPC formulation in handling whole-body disturbances while enforcing safety constraints.

Finally, we observe that the controller’s response varies depending on when in the gait cycle the disturbance occurs. Due to the robot's mass distribution, the rear leg joints experience higher loads than the front legs, making rear foot placement critical for disturbance recovery. At the first disturbance, the rear left foot is the next to step and requires minimal position adjustment, albeit pushing the calf torque to its \SI{320}{\newton\meter} limit. In contrast, during the second disturbance, the rear right foot must significantly modify its step location to prevent the base from drifting further left. This experiment highlights the MPC's ability to reason jointly about torques, footstep adaptation, and whole-body motion in response to external perturbations.

\subsubsection{With Time Delay}
\label{sec:res-sim-delay}
With the computation time delay accounted for in the simulator, our controller requires damping of about $\mathbf{K}_d = 5$ for stable walking and arm end-effector tracking. In this setting, we test the adaptive time step strategy described in~\Cref{sec:adap-steps}. We run the MPC with 15 nodes and adaptive steps ranging from \SI{10}{\milli\second} to \SI{100}{\milli\second}, totaling a horizon of \SI{0.56}{\second}. Comparing this with fixed time steps of \SI{40}{\milli\second}, resulting in the same horizon length, shows a significant reduction in drift of the base while trotting: \SI{0.4}{\centi\meter/\second} drift for adaptive time steps vs. \SI{1.2}{\centi\meter/\second} drift for fixed time steps.

\subsection{Hardware Experiments}
\label{res:hardware}

We deploy the MPC pipeline with the compiled \textit{Fatrop} solver on hardware. The controller runs at \SI{80}{\hertz} off-board, on the same processor as for the simulation experiments, and the control signals are sent over Ethernet to \textit{B2} and \textit{Z1}. \textit{B2} supports \textit{ROS2} for low-level communication, whereas \textit{Z1} provides a UDP-based Software Development Kit.

Unlike in simulation, where the controller performed well with pure feed-forward torques, hardware deployment required a velocity feedback gain of around $\boldsymbol{K}_d = 20$ to avoid high-frequency motor oscillations. Adding the position feedback gain $\boldsymbol{K}_p$ was not necessary for the quadruped, demonstrating the effectiveness of the MPC torque solution. For the manipulator, noisy joint sensor readings sometimes caused the end-effector position to deviate, which could be mitigated by adding a small proportional term for the arm joints.

\begin{figure}[t]
    \centering
    \begin{minipage}{\linewidth}
    \includegraphics[height=0.365\linewidth]{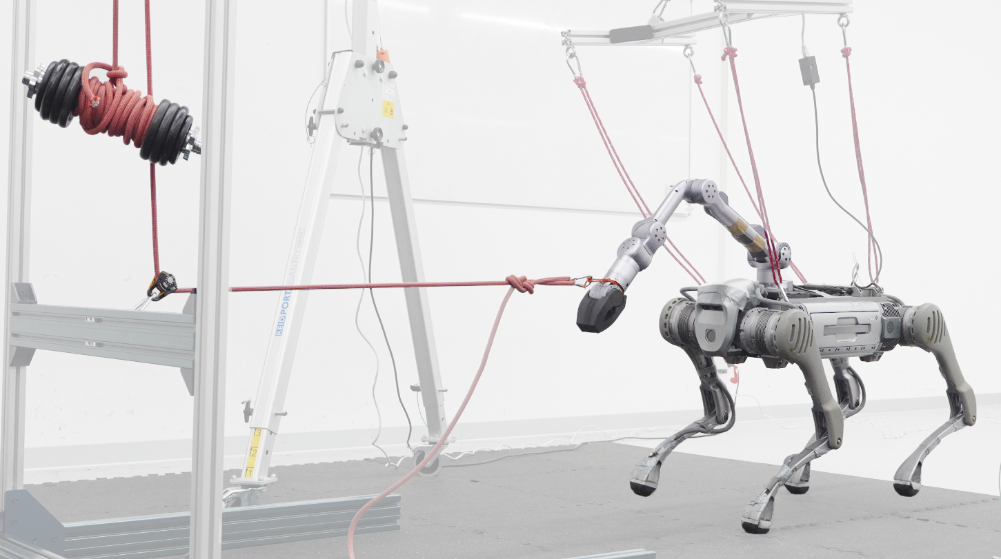}
    \hfill
    \includegraphics[height=0.365\linewidth]{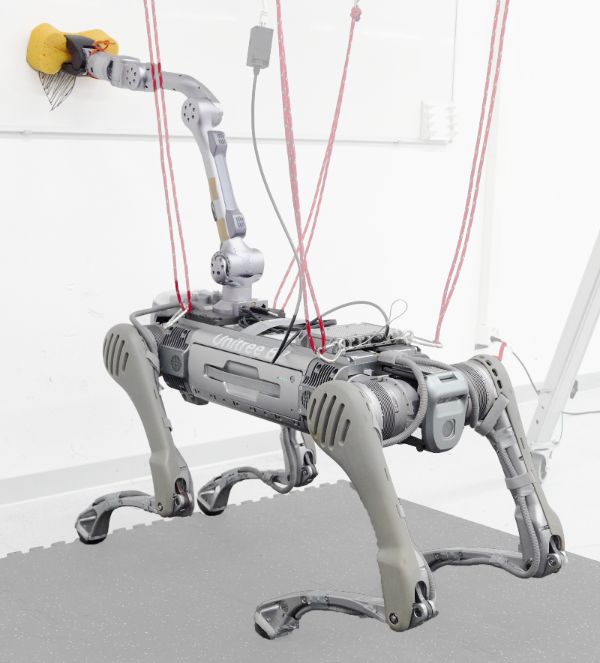}
    \end{minipage}
    \caption{Hardware demonstration of loco-manipulation tasks: Pulling a \SI{10}{\kilogram} load while walking (left) and accurately wiping a whiteboard (right).} 
    \label{fig:hw-demo2}
\end{figure}

We perform several whole-body loco-manipulation tasks on hardware (\Cref{fig:hw-demo1} and~\Cref{fig:hw-demo2}). To test the MPC near the robot's operational limits, we demonstrate pulling a \SI{10}{\kilogram} payload in stance and while walking. This is a significant increase over the Z1 arm's rated maximum payload of \SIrange{3}{5}{\kilogram}. Despite simulation showing successful pulling up to \SI{14}{\kilogram}, in a real scenario, overheating of the third arm joint prevents achieving higher loads.

Additional tasks include pushing a box against a wall while trotting in place (\Cref{fig:hw-demo1}), and wiping a whiteboard while standing (\Cref{fig:hw-demo2}). For these experiments, the end-effector force targets were manually specified. As we lacked the equipment to measure the actual applied forces, we could not verify whether the targets were precisely tracked. Nevertheless, the results demonstrate that our MPC can simultaneously handle end-effector force and motion targets on hardware.

Finally, by reducing the arm's PD gains and setting the end-effector target velocity and force to zero, we demonstrate compliant behavior for human interaction (\Cref{fig:hw-demo1}). This experiment indicates the potential for safe human-robot collaboration in real-world loco-manipulation scenarios.

\begin{figure}[t]
    \centering
    \begin{subfigure}{0.49\linewidth}
        \centering
        \includegraphics[width=\linewidth]{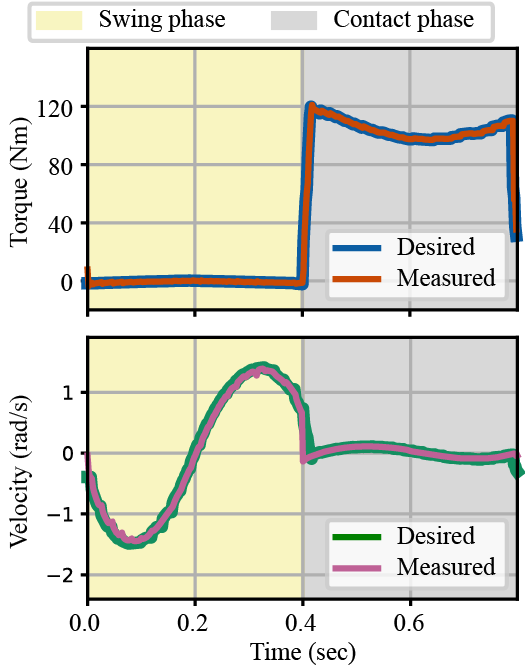}
        \caption{Simulation}
    \end{subfigure}
    \hfill
    \begin{subfigure}{0.49\linewidth}
        \centering
        \includegraphics[width=\linewidth]{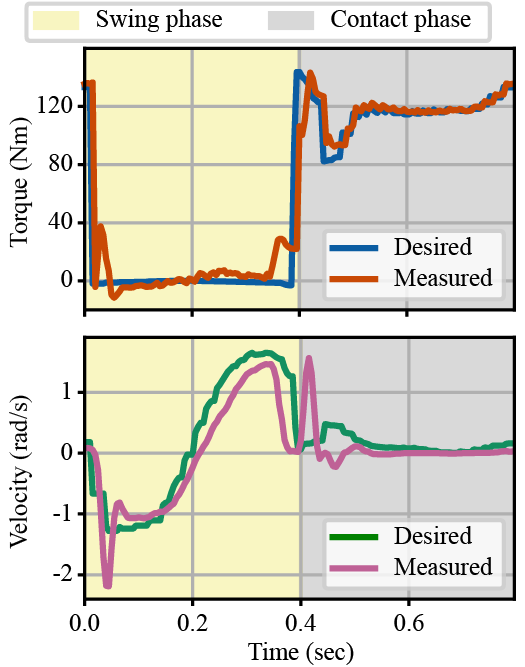}
        \caption{Hardware}
    \end{subfigure}
    \caption{Rear left calf joint torques and velocities during trot}
    \label{fig:data-hw-sim}
    \vspace{-0.2cm}
\end{figure}

\subsection{Sim-to-Real Discrepancies}
Despite the successful hardware deployment, the experiments revealed several sim-to-real mismatches that affect real-world performance. Most notably, solve times showed greater variability on hardware, occasionally reaching up to \SI{20}{\milli\second} compared to an average of \SI{12.5}{\milli\second} in simulation. This may stem from computation delays, causing the MPC prediction to deviate from the true system state, or from model discrepancies. 

To investigate the sim-to-real mismatch, we compare joint-level data from the rear-left calf on hardware and in simulation (\Cref{fig:data-hw-sim}). Notable differences appear in both torque and velocity profiles, particularly during contact transitions. At the start of the contact phase, the calf torque spikes to support the robot’s weight. In simulation, this transition is stable for both torque and velocity. On hardware, however, a spike in joint velocity indicates delayed ground contact, prompting the MPC to reduce torque until contact is established. This contact mode mismatch is a well-known challenge in model-based control of legged robots.

Beyond contact transitions, other discrepancies also affect performance. First, the dynamics model does not account for motor friction and rotor inertia, which degrade torque tracking on hardware.
Prior work addresses this by incorporating these effects directly into the MPC model~\citep{mit-humanoid} or compensating via feed-forward torque terms~\citep{wb-talos, wb-humanoid}.
Applying similar techniques to the \textit{B2} and \textit{Z1} platforms, based on system identification~\citep{torque-sysid}, could improve tracking accuracy.
Second, we observed that the robot's mass distribution deviates from the URDF: While the model places the base center of mass within \SI{1}{\centi\meter} of its geometric center, hardware measurements indicated a rearward weight shift of about \SI{60}{\percent}. Updating the CoM offset and contact force targets in the MPC significantly reduced backward lean on hardware, and all results reported here were obtained with this modification.
This finding suggests that more advanced system identification methods~\citep{figaroh} could further improve model accuracy and controller performance.

\section{Conclusion and Future Work}

We present a real-time whole-body MPC framework for loco-manipulation on the \textit{Unitree B2} quadruped equipped with a \textit{Z1} manipulator. Our method directly optimizes joint torques by enforcing whole-body inverse dynamics as path constraints on each node. Using the interior-point solver \textit{Fatrop}, we successfully transfer the MPC to hardware, running the optimization at \SI{80}{\hertz} and interpolating the solution at \SI{500}{\hertz} to interface with a low-level PD controller. Crucially, no separate whole-body controller is required to track the MPC solution, resulting in a simple unified pipeline. 

Future work should focus on extending the framework to model object-level dynamics for interaction~\citep{unified-mpc, versatile-locoman}, allowing the MPC to optimize arm end-effector force and motion instead of having them as an external command. Additionally, collision constraints should be incorporated for both the quadruped and the manipulator to ensure safety. A thorough evaluation of alternative solution methods, such as sequential quadratic programming, is also planned, given its promising performance in real-time MPC for locomotion~\citep{perceptive-loco, mit-humanoid}. Finally, to address the sim-to-real mismatches observed in our hardware experiments, system identification methods should be employed to improve low-level torque tracking and refine the robot’s inertial properties.

The sim-to-real limitations of MPC under model inaccuracies also motivate reinforcement learning (RL) as a complementary approach. Through domain randomization~\citep{domain-rand}, RL has shown robustness to modeling errors in real-world deployments, and recent work demonstrates its potential for loco-manipulation~\citep{learning-compliance, rsl-locoman}. However, exploration in high-dimensional systems remains challenging, and learned policies typically operate at the position level---limiting their ability to plan both motion and force. Integrating RL with MPC is a promising direction for combining learning-based robustness with model-based planning and safety~\citep{kang2023rl+, deep-tracking-control, cheng2025rambo}.


\section*{Acknowledgments}
{
The authors thank Charles Khazoom for discussing the whole-body MPC on the MIT humanoid \citep{mit-humanoid}, Jakob Genhart for implementing the Z1 arm communication interface, and Ajay Sathya for his insights on warm-starting \textit{Fatrop}.
}


\bibliographystyle{IEEEtranN}
\bibliography{root.bib}

@inproceedings{pinocchio,
  title={The Pinocchio C++ library: A fast and flexible implementation of rigid body dynamics algorithms and their analytical derivatives},
  author={Carpentier, Justin and Saurel, Guilhem and Buondonno, Gabriele and Mirabel, Joseph and Lamiraux, Florent and Stasse, Olivier and Mansard, Nicolas},
  booktitle={2019 IEEE/SICE International Symposium on System Integration (SII)},
  pages={614--619},
  year={2019},
  organization={IEEE}
}

@misc{aligator,
  author = {Jallet, Wilson and Bambade, Antoine and El Kazdadi, Sarah and Justin, Carpentier and Nicolas, Mansard},
  title = {aligator},
  url = {https://github.com/Simple-Robotics/aligator}
}

@Article{casadi,
  author = {Joel A E Andersson and Joris Gillis and Greg Horn
            and James B Rawlings and Moritz Diehl},
  title = {{CasADi} -- {A} software framework for nonlinear optimization
           and optimal control},
  journal = {Mathematical Programming Computation},
  volume = {11},
  number = {1},
  pages = {1--36},
  year = {2019},
  publisher = {Springer},
  
}

@ARTICLE{unified-mpc,
  author={Sleiman, Jean-Pierre and Farshidian, Farbod and Minniti, Maria Vittoria and Hutter, Marco},
  journal={IEEE Robotics and Automation Letters}, 
  title={A Unified MPC Framework for Whole-Body Dynamic Locomotion and Manipulation}, 
  year={2021},
  volume={6},
  number={3},
  pages={4688-4695},
  }

@ARTICLE{mit-humanoid,
  author={Khazoom, Charles and Hong, Seungwoo and Chignoli, Matthew and Stanger-Jones, Elijah and Kim, Sangbae},
  journal={IEEE Robotics and Automation Letters}, 
  title={Tailoring Solution Accuracy for Fast Whole-Body Model Predictive Control of Legged Robots}, 
  year={2024},
  volume={9},
  number={12},
  pages={11074-11081},
  }

@inproceedings{fatrop,
  title={Fatrop: A fast constrained optimal control problem solver for robot trajectory optimization and control},
  author={Vanroye, Lander and Sathya, Ajay and De Schutter, Joris and Decr{\'e}, Wilm},
  booktitle={2023 IEEE/RSJ International Conference on Intelligent Robots and Systems (IROS)},
  pages={10036--10043},
  year={2023},
  organization={IEEE}
}

@INPROCEEDINGS{mit-cheetah,
  author={Di Carlo, Jared and Wensing, Patrick M. and Katz, Benjamin and Bledt, Gerardo and Kim, Sangbae},
  booktitle={2018 IEEE/RSJ International Conference on Intelligent Robots and Systems (IROS)}, 
  title={Dynamic Locomotion in the MIT Cheetah 3 Through Convex Model-Predictive Control}, 
  year={2018},
  volume={},
  number={},
  pages={1-9},
  }

@article{mit-cheetah-wbc,
  title={Highly dynamic quadruped locomotion via whole-body impulse control and model predictive control},
  author={Kim, Donghyun and Di Carlo, Jared and Katz, Benjamin and Bledt, Gerardo and Kim, Sangbae},
  journal={arXiv preprint arXiv:1909.06586},
  year={2019}
}

@article{humanoid-locoman,
  title={Kinodynamics-based pose optimization for humanoid loco-manipulation},
  author={Li, Junheng and Nguyen, Quan},
  journal={arXiv preprint arXiv:2303.04985},
  year={2023}
}

@inproceedings{hierarchical-locoman,
  title={Hierarchical optimization-based control for whole-body loco-manipulation of heavy objects},
  author={Rigo, Alberto and Hu, Muqun and Gupta, Satyandra K and Nguyen, Quan},
  booktitle={2024 IEEE International Conference on Robotics and Automation (ICRA)},
  pages={15322--15328},
  year={2024},
  organization={IEEE}
}

@INPROCEEDINGS{centroidal-dyn-full-kin,
  author={Dai, Hongkai and Valenzuela, Andrés and Tedrake, Russ},
  booktitle={IEEE-RAS International Conference on Humanoid Robots}, 
  title={Whole-body motion planning with centroidal dynamics and full kinematics}, 
  year={2014},
  volume={},
  number={},
  pages={295-302},
  }

@ARTICLE{inv-dyn-mpc,
  author={Mastalli, Carlos and Chhatoi, Saroj Prasad and Corbéres, Thomas and Tonneau, Steve and Vijayakumar, Sethu},
  journal={IEEE Transactions on Robotics}, 
  title={Inverse-Dynamics MPC via Nullspace Resolution}, 
  year={2023},
  volume={39},
  number={4},
  pages={3222-3241},
  }

@article{versatile-locoman,
  title={Versatile multicontact planning and control for legged loco-manipulation},
  author={Sleiman, Jean-Pierre and Farshidian, Farbod and Hutter, Marco},
  journal={Science Robotics},
  volume={8},
  number={81},
  pages={eadg5014},
  year={2023},
  publisher={American Association for the Advancement of Science}
}

@article{perceptive-loco,
  title={Perceptive locomotion through nonlinear model-predictive control},
  author={Grandia, Ruben and Jenelten, Fabian and Yang, Shaohui and Farshidian, Farbod and Hutter, Marco},
  journal={IEEE Transactions on Robotics},
  volume={39},
  number={5},
  pages={3402--3421},
  year={2023},
  publisher={IEEE}
}

@INPROCEEDINGS{centroidal-37dof,
  author={Dadiotis, Ioannis and Laurenzi, Arturo and Tsagarakis, Nikos},
  booktitle={2023 IEEE-RAS 22nd International Conference on Humanoid Robots (Humanoids)}, 
  title={Whole-Body MPC for Highly Redundant Legged Manipulators: Experimental Evaluation with a 37 DoF Dual-Arm Quadruped}, 
  year={2023},
  volume={},
  number={},
  pages={1-8},
  }

@INPROCEEDINGS{centroidal-wb-comparison,
  author={Dantec, Ewen and Jallet, Wilson and Carpentier, Justin},
  booktitle={2024 IEEE-RAS 23rd International Conference on Humanoid Robots (Humanoids)}, 
  title={From centroidal to whole-body models for legged locomotion: a comparative analysis}, 
  year={2024},
  volume={},
  number={},
  pages={335-342},
  }

@INPROCEEDINGS{wb-talos,
  author={Dantec, Ewen and Budhiraja, Rohan and Roig, Adria and Lembono, Teguh and Saurel, Guilhem and Stasse, Olivier and Fernbach, Pierre and Tonneau, Steve and Vijayakumar, Sethu and Calinon, Sylvain and Taix, Michel and Mansard, Nicolas},
  booktitle={2021 IEEE International Conference on Robotics and Automation (ICRA)}, 
  title={Whole Body Model Predictive Control with a Memory of Motion: Experiments on a Torque-Controlled Talos}, 
  year={2021},
  volume={},
  number={},
  pages={8202-8208},
  }

@unpublished{wb-impedance,
  TITLE = {{Whole-Body MPC without Foot References for the Locomotion of an Impedance-Controlled Robot}},
  AUTHOR = {Assirelli, Alessandro and Risbourg, Fanny and Lunardi, Gianni and Flayols, Thomas and Mansard, Nicolas},
  NOTE = {working paper or preprint},
  YEAR = {2022},
}

@INPROCEEDINGS{wb-humanoid,
  author={Dantec, Ewen and Naveau, Maximilien and Fernbach, Pierre and Villa, Nahuel and Saurel, Guilhem and Stasse, Olivier and Taix, Michel and Mansard, Nicolas},
  booktitle={2022 IEEE-RAS 21st International Conference on Humanoid Robots (Humanoids)}, 
  title={Whole-Body Model Predictive Control for Biped Locomotion on a Torque-Controlled Humanoid Robot}, 
  year={2022},
  volume={},
  number={},
  pages={638-644},
  }

@article{ipopt,
  title={On the Implementation of a Primal-Dual Interior Point Filter Line Search Algorithm for Large-Scale Nonlinear Programming},
  author={W{\"a}chter, A. and Biegler, L. T.},
  journal={Mathematical Programming},
  volume={106},
  number={1},
  pages={25--57},
  year={2006},
}

@book{rigid-body-dynamics,
  title={Rigid Body Dynamics Algorithms},
  author={Featherstone, Roy},
  year={2007},
  publisher={Springer-Verlag},
  address={Berlin, Heidelberg}
}

@INPROCEEDINGS{domain-rand,
  author={Tobin, Josh and Fong, Rachel and Ray, Alex and Schneider, Jonas and Zaremba, Wojciech and Abbeel, Pieter},
  booktitle={2017 IEEE/RSJ International Conference on Intelligent Robots and Systems (IROS)}, 
  title={Domain randomization for transferring deep neural networks from simulation to the real world}, 
  year={2017},
  volume={},
  number={},
  pages={23-30},
  }

@inproceedings{rsl-locoman,
  title={Learning to Open and Traverse Doors with a Legged Manipulator},
  author={Zhang, Mike and Ma, Yuntao and Miki, Takahiro and Hutter, Marco},
  booktitle={Proceedings of The 8th Conference on Robot Learning},
  series={Proceedings of Machine Learning Research},
  volume={270},
  pages={2913--2927},
  year={2025},
  publisher={PMLR}
}

@INPROCEEDINGS{learning-compliance,
  author={Portela, Tifanny and Margolis, Gabriel B. and Ji, Yandong and Agrawal, Pulkit},
  booktitle={2024 IEEE International Conference on Robotics and Automation (ICRA)}, 
  title={Learning Force Control for Legged Manipulation}, 
  year={2024},
  volume={},
  number={},
  pages={15366-15372},
  }

@article{deep-tracking-control,
  title={Dtc: Deep tracking control},
  author={Jenelten, Fabian and He, Junzhe and Farshidian, Farbod and Hutter, Marco},
  journal={Science Robotics},
  volume={9},
  number={86},
  pages={eadh5401},
  year={2024},
  publisher={American Association for the Advancement of Science}
}

@article{rsl-dynamic-locomotion,
  title={Dynamic locomotion through online nonlinear motion optimization for quadrupedal robots},
  author={Bellicoso, C Dario and Jenelten, Fabian and Gehring, Christian and Hutter, Marco},
  journal={IEEE Robotics and Automation Letters},
  volume={3},
  number={3},
  pages={2261--2268},
  year={2018},
  publisher={IEEE}
}

@article{rsl-wb-mpc,
  title={Whole-body nonlinear model predictive control through contacts for quadrupeds},
  author={Neunert, Michael and St{\"a}uble, Markus and Giftthaler, Markus and Bellicoso, Carmine D and Carius, Jan and Gehring, Christian and Hutter, Marco and Buchli, Jonas},
  journal={IEEE Robotics and Automation Letters},
  volume={3},
  number={3},
  pages={1458--1465},
  year={2018},
  publisher={IEEE}
}

@article{ode,
  title={Open Dynamics Engine},
  author={Smith, Russell and others},
  journal={http://www.ode.org},
  year={2005},
  note={Accessed: 2025-05-30}
}

@article{fatrop-riccati,
  title={A generalization of the Riccati recursion for equality-constrained linear quadratic optimal control},
  author={Vanroye, Lander and De Schutter, Joris and Decr{\'e}, Wilm},
  journal={Optimal Control Applications and Methods},
  volume={45},
  number={1},
  pages={436--454},
  year={2024},
  publisher={Wiley Online Library}
}

@article{anymal-srb,
  title={Gait and trajectory optimization for legged systems through phase-based end-effector parameterization},
  author={Winkler, Alexander W and Bellicoso, C Dario and Hutter, Marco and Buchli, Jonas},
  journal={IEEE Robotics and Automation Letters},
  volume={3},
  number={3},
  pages={1560--1567},
  year={2018},
  publisher={IEEE}
}

@inproceedings{constrained-srb,
  title={Real-time constrained nonlinear model predictive control on so (3) for dynamic legged locomotion},
  author={Hong, Seungwoo and Kim, Joon-Ha and Park, Hae-Won},
  booktitle={2020 IEEE/RSJ International Conference on Intelligent Robots and Systems (IROS)},
  pages={3982--3989},
  year={2020},
  organization={IEEE}
}

@inproceedings{robust-wbc,
  title={Computationally-robust and efficient prioritized whole-body controller with contact constraints},
  author={Kim, Donghyun and Lee, Jaemin and Ahn, Junhyeok and Campbell, Orion and Hwang, Hochul and Sentis, Luis},
  booktitle={IEEE/RSJ International Conference on Intelligent Robots and Systems (IROS)},
  pages={1--8},
  year={2018},
  organization={IEEE}
}

@inproceedings{figaroh,
  title={FIGAROH: a Python toolbox for dynamic identification and geometric calibration of robots and humans},
  author={Nguyen, Thanh DV and Bonnet, Vincent and Sabbah, Maxime and Gautier, Maxime and Fernbach, Pierre and Lamiraux, Florent},
  booktitle={2023 IEEE-RAS 22nd International Conference on Humanoid Robots (Humanoids)},
  pages={1--8},
  year={2023},
  organization={IEEE}
}

@inproceedings{mit-cheetah-3,
  title={Mit cheetah 3: Design and control of a robust, dynamic quadruped robot},
  author={Bledt, Gerardo and Powell, Matthew J and Katz, Benjamin and Di Carlo, Jared and Wensing, Patrick M and Kim, Sangbae},
  booktitle={2018 IEEE/RSJ International Conference on Intelligent Robots and Systems (IROS)},
  pages={2245--2252},
  year={2018},
  organization={IEEE}
}

@inproceedings{torque-sysid,
  title={Improving Domain Transfer of Robot Dynamics Models with Geometric System Identification and Learned Friction Compensation},
  author={Schwendeman, Laura and SaLoutos, Andrew and Stanger-Jones, Elijah and Kim, Sangbae},
  booktitle={2023 IEEE-RAS 22nd International Conference on Humanoid Robots (Humanoids)},
  pages={1--8},
  year={2023},
  organization={IEEE}
}

@article{wb-mobile-manipulator,
  title={Whole-body mpc for a dynamically stable mobile manipulator},
  author={Minniti, Maria Vittoria and Farshidian, Farbod and Grandia, Ruben and Hutter, Marco},
  journal={IEEE Robotics and Automation Letters},
  volume={4},
  number={4},
  pages={3687--3694},
  year={2019},
  publisher={IEEE}
}

@article{diffusing-horizon-mpc,
  title={Diffusing-horizon model predictive control},
  author={Shin, Sungho and Zavala, Victor M},
  journal={IEEE Transactions on Automatic Control},
  volume={68},
  number={1},
  pages={188--201},
  year={2021},
  publisher={IEEE}
}

@article{cafe-mpc,
  title={Cafe-mpc: A cascaded-fidelity model predictive control framework with tuning-free whole-body control},
  author={Li, He and Wensing, Patrick M},
  journal={IEEE Transactions on Robotics},
  year={2024},
  publisher={IEEE}
}

@article{roloma,
  title={Roloma: Robust loco-manipulation for quadruped robots with arms},
  author={Ferrolho, Henrique and Ivan, Vladimir and Merkt, Wolfgang and Havoutis, Ioannis and Vijayakumar, Sethu},
  journal={Autonomous Robots},
  volume={47},
  number={8},
  pages={1463--1481},
  year={2023},
  publisher={Springer}
}

@inproceedings{inv-vs-forw-dyn,
  title={Inverse dynamics vs. forward dynamics in direct transcription formulations for trajectory optimization},
  author={Ferrolho, Henrique and Ivan, Vladimir and Merkt, Wolfgang and Havoutis, Ioannis and Vijayakumar, Sethu},
  booktitle={2021 IEEE International Conference on Robotics and Automation (ICRA)},
  pages={12752--12758},
  year={2021},
  organization={IEEE}
}

@article{kang2023rl+,
  title={Rl+ model-based control: Using on-demand optimal control to learn versatile legged locomotion},
  author={Kang, Dongho and Cheng, Jin and Zamora, Miguel and Zargarbashi, Fatemeh and Coros, Stelian},
  journal={IEEE Robotics and Automation Letters},
  volume={8},
  number={10},
  pages={6619--6626},
  year={2023},
  publisher={IEEE}
}

@article{cheng2025rambo,
  title={RAMBO: RL-augmented Model-based Optimal Control for Whole-body Loco-manipulation},
  author={Cheng, Jin and Kang, Dongho and Fadini, Gabriele and Shi, Guanya and Coros, Stelian},
  journal={arXiv preprint arXiv:2504.06662},
  year={2025}
}

@inproceedings{cheng2022haptic,
  title={Haptic teleoperation of high-dimensional robotic systems using a feedback mpc framework},
  author={Cheng, Jin and Abi-Farraj, Firas and Farshidian, Farbod and Hutter, Marco},
  booktitle={2022 IEEE/RSJ International Conference on Intelligent Robots and Systems (IROS)},
  pages={6197--6204},
  year={2022},
  organization={IEEE}
}

\end{document}